\documentclass[lettersize,journal]{IEEEtran}
\usepackage{amsmath,amsfonts}
\usepackage{algorithmic}
\usepackage{algorithm}
\usepackage{array}
\usepackage[caption=false,font=normalsize,labelfont=sf,textfont=sf]{subfig}
\usepackage{textcomp}
\usepackage{stfloats}
\usepackage{url}
\usepackage{verbatim}
\usepackage{graphicx}
\usepackage{bm}
\usepackage{xcolor}
\usepackage[numbers,sort&compress]{natbib}
\usepackage{booktabs}
\usepackage{multirow}
\usepackage{adjustbox}
\usepackage{orcidlink}
\hypersetup{hidelinks}
\usepackage{float}
\usepackage{makecell}
\usepackage{booktabs}
\usepackage{diagbox}
\usepackage{lineno}
\newcolumntype{C}[1]{>{\centering\arraybackslash}p{#1}}
\begin{document}

\title{{\fontsize{21pt}{30pt}\selectfont
DisasterTD: Disaster Toponym Disambiguation Using Multimodal LLMs and Cross-View Geolocalization}}

\author{
Wenping Yin\textsuperscript{\orcidlink{0009-0002-2056-0091}},
Ziqi Liu\textsuperscript{\orcidlink{0009-0001-0483-5502}},
Naixia Mou\textsuperscript{\orcidlink{0000-0003-1700-5943}},
Weijia Li\textsuperscript{\orcidlink{0000-0003-1838-9176}},~\IEEEmembership{Member,~IEEE},
Danfeng Hong\textsuperscript{\orcidlink{0000-0002-3212-9584}},~\IEEEmembership{Senior Member,~IEEE}, \\
Hao Li\textsuperscript{\orcidlink{0000-0002-6336-8772}},~\IEEEmembership{Member,~IEEE}

\thanks{This is the accepted version of a paper to appear in IEEE Transactions on Geoscience and Remote Sensing.}
\thanks{This work was partly supported by the Start-Up Grant (SUG) project ``Geospatial Artificial Intelligence for Climate Resilient Urban Environment" from the National University of Singapore. \textit{(Corresponding author: Hao Li)}}
\thanks{Wenping Yin is with the College of Geodesy and Geomatics, Shandong University of Science and Technology, Qingdao 266590, China, and also with the School of Environmental Science and Spatial Informatics, China University of Mining and Technology, Xuzhou 221116, China (email: yin@cumt.edu.cn).}
\thanks{Ziqi Liu is with the State Key Laboratory of Information Engineering in Surveying, Mapping and Remote Sensing, Wuhan University, Wuhan 430079, China (email: lzq677@whu.edu.cn).}
\thanks{Naixia Mou is with the College of Geodesy and Geomatics, Shandong University of Science and Technology, Qingdao 266590, China (email: mounaixia@163.com).}
\thanks{Weijia Li is with the Tsinghua Shenzhen International Graduate School, Tsinghua University, Shenzhen 518055, China. (email: liweijia@sz.tsinghua.edu.cn).}
\thanks{Danfeng Hong is with the School of Automation, Southeast University, Nanjing 210096, China (email:  hongdanfeng1989@gmail.com).}
\thanks{Hao Li is with the Department of Geography, National University of Singapore, Singapore 117568, Singapore (email: hao.li@nus.edu.sg).}
}

\markboth{
IEEE Transactions on Geoscience and Remote Sensing, 2026}
{Shell \MakeLowercase{\textit{et al.}} A Sample Article Using IEEEtran.cls for IEEE Journals}


\maketitle


\begin{abstract}
Social media imagery (SMI) provides timely and fine-grained ground perspectives that are valuable for situational awareness and emergency response. Unlike satellite or aerial imagery, SMI can capture disaster impacts and ground-level conditions in a timely manner. However, geographic references in SMI are often vague or ambiguous, making accurate geolocalization challenging. To address this issue, we propose DisasterTD, a disaster toponym disambiguation framework that integrates multimodal large language model (MLLMs)-based semantic reasoning with cross-view geolocalization. First, MLLMs extract toponyms and generate candidate geolocations from noisy textual inputs. Then, cross-view matching between SMI, remote sensing imagery (RSI), and optionally street-view imagery (SVI) is used to verify and refine these candidate results. A Vision Transformer (ViT)-based visual foundation model, DINOv2, is used to bridge the domain gap between overhead and ground-level imagery. We evaluate DisasterTD on the Hurricane Harvey dataset, where SMI is augmented with collected RSI and SVI to construct a cross-view benchmark for disaster geolocalization. The dataset is divided into four categories based on toponym clarity and ambiguity, allowing a fine-grained performance analysis across scenarios. Results show that DisasterTD consistently outperforms MLLM-only and cross-view-only baselines without disambiguation, achieving geolocalization accuracies of 71.62\% within 1000 m, 62.36\% within 500 m, 57.99\% within 250 m, 52.09\% within 100 m, and 47.01\% within 50 m, while reducing the mean and median errors to 11.33 km and 0.68 km, respectively. The largest improvements appear in ambiguous toponyms, where semantic reasoning with cross-view evidence reduces candidate dispersion and errors. These findings demonstrate the effectiveness of integrating MLLM-based candidate generation with cross-view verification for fine-grained disaster geolocalization.
\end{abstract}



\section{Introduction}

\IEEEPARstart{N}{atural} disasters such as hurricanes, floods, wildfires, and earthquakes have become increasingly frequent and severe in recent decades, posing growing threats to human lives, infrastructure, and ecosystems  \citep{dietze2021flood,nohrstedt2022exploring, shirmohammadi2026challenges}. Rapid and accurate geolocalization of disaster-related information is crucial for effective emergency response, resource allocation, and situational awareness \citep{li2025cross,yin2025triple}. In particular, social media imagery (SMI) has emerged as a valuable and increasingly important complement to traditional remote sensing data, providing timely on-the-ground perspectives that reveal details unavailable in satellite or aerial observations \citep{feng2022extraction,hu2023location}. Unlike satellite or aerial imagery, which may suffer from latency or limited coverage, SMIs are generated rapidly by affected populations and can capture fine-grained, ground-level impacts. Furthermore, these images often include contextual cues such as damaged infrastructure, water levels, or fire spread patterns, which are highly informative for rapid damage assessment. However, fully leveraging these crowdsourced data requires extracting reliable geographic information, making robust geolocalization essential for modern disaster management.

To address this need, researchers have explored various methods for text- and image-based geolocalization. Textual approaches typically extract explicit or implicit geographic references from unstructured messages and map them to candidate geolocations \citep{hu2023geo,huang2023survey}. These methods often rely on named entity recognition, gazetteer matching, or contextual language models to identify place mentions, but they can struggle when information is vague or incomplete. Visual methods rely on matching image content to reference databases such as remote sensing imagery (RSI) or street-view imagery (SVI) \citep{zhuang2021faster, deuser2023sample4geo, gong2024satellite, hong2026hyperspectral}. Common image-based methods include ground-view image retrieval, cross-view image retrieval, and two-dimensional–three-dimensional (2D–3D) matching, each designed to handle different types of image data \citep{huang2023survey, fang2025scof, li2025unsupervised, liang2025dstg}. Recent advances in natural language processing (NLP) and computer vision (CV), especially with the advent of multimodal large language models (MLLMs) and deep neural networks, have significantly improved the capability of extracting and aligning geographic cues across modalities \citep{yin2025llm, hong2026foundation}. Despite these developments, important challenges remain: textual cues can be vague or ambiguous, while visual data are often affected by occlusion, scene complexity, or domain gaps between ground-level and overhead imagery. Text-only and image-only strategies face significant limitations in disaster geolocalization, where information is often noisy, incomplete, and highly time-sensitive.

One major challenge in this context lies in toponym disambiguation \citep{hu2023can,li2026towards}. Toponyms in SMIs often correspond to multiple geolocations (e.g., common landmarks, chain stores, or generic street signs), making it difficult to directly link them to unique geographic coordinates \citep{grace2021toponym}. This ambiguity is further exacerbated in disaster scenarios, where users frequently provide incomplete, colloquial, or misspelled place references. Traditional disambiguation approaches, such as gazetteer lookups, probabilistic models, or contextual cues from surrounding text, often perform poorly when confronted with the short, informal, and noisy nature of user-generated content on social platforms \citep{middleton2018location,fize2021deep}. More recent efforts attempt to mitigate these issues by combining semantic reasoning with spatial constraints, user metadata, visual grounding, and integrating MLLMs with geo-knowledge \citep{hu2024toponym,wang2020neurotpr}. While these strategies demonstrate improvements, they remain sensitive to data sparsity and domain shifts, and a robust solution capable of handling unambiguous and ambiguous toponyms in complex disaster settings is still lacking.

\begin{figure}[htbp]
\centering
\includegraphics[width=\linewidth]{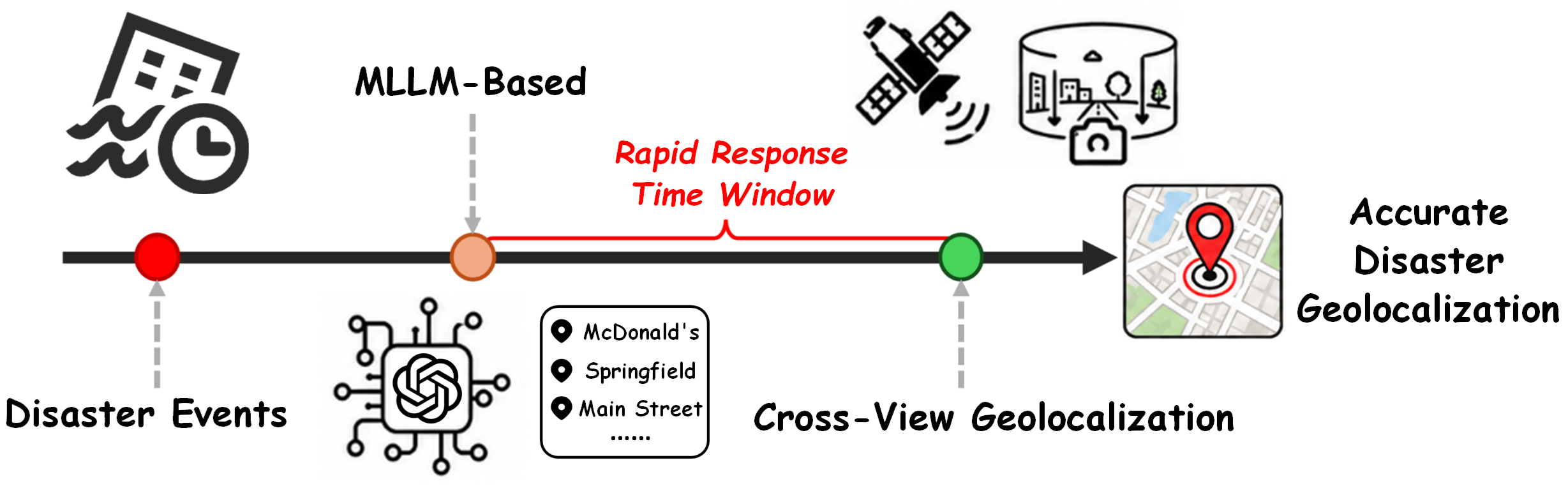}
\vspace{-10pt}
\caption{A conceptual diagram of DisasterTD for disaster geolocalization, illustrating how MLLM-based toponym extraction and cross-view geolocalization enable toponym disambiguation and accurate geolocalization within a critical response time window.}
\label{ConceptualDiagram}
\end{figure}

\textcolor{black}{In this study, we propose DisasterTD, a novel remote sensing–based framework for disaster toponym disambiguation that integrates MLLMs with cross-view geolocalization. As shown in Fig. \ref{ConceptualDiagram}, DisasterTD first uses MLLMs to extract toponyms and generate candidate geolocations, followed by cross-view visual matching between SMIs, RSIs, and optionally SVIs to refine and verify the results. By combining semantic reasoning with multi-source Earth observation data, DisasterTD addresses the limitations of text-only and image-only strategies in complex disaster geolocalization scenarios. Unlike our previous MLLM-based disaster geolocalization work \citep{yin2025llm}, which mainly used extracted geoinformation as direct geolocalization cues, DisasterTD treats MLLM-generated geolocations as candidate geolocations that require further disambiguation. By integrating semantic candidate generation with cross-view visual verification, DisasterTD provides a new dedicated solution for disaster toponym disambiguation.}

The remainder of this paper is organized as follows: Section 2 reviews related work on geolocalization and toponym disambiguation. Section 3 details the proposed DisasterTD, including MLLM-based initial disaster geolocalization and cross-view matching disambiguation strategies. Section 4 presents experimental settings and results on the extended Hurricane Harvey dataset. Section 5 discusses the findings and limitations, and Section 6 concludes with future research directions and broader GeoAI and remote sensing applications.

\section{Related Work}
\subsection{Remote Sensing Assisted Geolocalization}

RSI serves as a valuable foundation for large-scale geolocalization by providing consistent and up-to-date overhead perspectives \citep{li2020integrating, hong2023cross}. Early studies relied heavily on template matching or hand-crafted feature descriptors such as scale-invariant feature transform (SIFT) and speeded-up robust features (SURF) to align ground-level photographs or SMI with corresponding regions in remote sensing datasets \citep{chen2011city, hays2008im2gps}. While these methods demonstrated the feasibility of leveraging RSI for geolocalization, their effectiveness was often limited by variations in scale, illumination, and seasonal conditions. The development of high-resolution satellite imagery and the availability of large-scale benchmarks have since enabled more sophisticated research on remote sensing assisted geolocalization \citep{zhai2017predicting,yin2025triple}. For example, \citet{workman2015wide} proposed wide-area geolocalization using aerial reference imagery, showing that global coverage can significantly expand applicability, while Vo and Hays \citep{vo2016localizing} demonstrated how overhead imagery can be used to localize and orient street-view photographs.

The rise of deep learning has brought various neural network architectures to the forefront of remote sensing–based geolocalization \citep{deuser2023sample4geo,li2025cross}. These models can extract high-level semantic representations from imagery, which are significantly more robust to variations in viewpoint, illumination, and environmental conditions compared with traditional feature-based methods. Cross-view geolocalization has emerged as a particularly effective strategy, where ground-level images or SVI are matched against overhead RSI to bridge perspective differences \citep{shi2020optimal,zheng2020university}. Such methods have demonstrated promising applications in urban mapping, navigation, and fine-grained geolocation recognition, demonstrating the potential of integrating RSI for fine-grained geolocalization \citep{fang2025scof, li2025unsupervised, liang2025dstg}. However, significant challenges remain, including occlusions, complex scene clutter, and the inherent domain gap between ground-level and aerial perspectives, which can limit matching accuracy in heterogeneous environments.

More recent studies have explored multimodal frameworks that integrate RSI with additional complementary signals, such as textual metadata, sensor readings, or social media streams \citep{wu2022im2city,yin2025llm}. By exploiting these heterogeneous information sources, such methods enhance robustness to noise, missing data, and environmental variations, thereby improving geolocalization accuracy in complex, real-world scenarios. \citet{shi2019spatial}, for example, introduced a spatial-aware feature aggregation mechanism that encodes geometric relationships, leading to more discriminative cross-view feature matching. In disaster contexts, remote sensing assisted geolocalization has proven particularly valuable, providing a scalable way to cross-validate ground-level observations against overhead evidence \citep{kustu2023deep,li2025cross}. However, the problem of ambiguity and incomplete cues in disaster imagery persists, motivating the need for integrated frameworks that combine remote sensing with semantic reasoning and cross-view disambiguation.

\subsection{Text- or Image-Based Geolocalization}

Text-based geolocalization has been widely studied in NLP and geographic information science, especially within the context of social media, news articles, and disaster reports where textual descriptions often contain explicit or implicit geographic references \citep{leidner2007toponym,lourentzou2017text}. Early approaches primarily relied on gazetteer matching and rule-based heuristics, mapping extracted toponyms to candidate geolocations through string similarity and hierarchical geographic filters \citep{amitay2004web}. Probabilistic models and topic-based methods were later introduced to integrate contextual signals such as co-occurring toponyms, temporal patterns, or linguistic cues, thereby improving disambiguation under sparse or noisy data \citep{wing2011simple,eisenstein2010latent}. More recently, neural architectures, particularly transformer-based models such as bidirectional encoder representations from transformers (BERT) and generative pre-trained transformer (GPT), have greatly advanced this field by capturing semantic nuances, handling informal and ambiguous expressions, and leveraging pretrained knowledge to support robust inference \citep{kamalloo2018coherent,hu2024toponym,bicakci2025street}. Despite these advances, challenges remain in resolving colloquial place mentions, handling spelling variations, and disambiguating vague toponyms without strong geographic anchors.

Image-based geolocalization aims to determine the geolocation of a photo or RSI by analyzing its visual content. Traditional methods employed handcrafted descriptors such as SIFT or SURF to detect distinctive keypoints and match them against geo-referenced databases \citep{lowe2004distinctive,bay2006surf}. With the rise of deep learning, convolutional neural networks (CNNs) and vision transformers (ViTs) have become the dominant paradigm, enabling end-to-end feature learning and large-scale image retrieval for geolocalization \citep{weyand2016planet,yi2025geolocsft}. These approaches have achieved remarkable progress in tasks ranging from street-view to remote-sensing image alignment, exploiting architectural structures, vegetation patterns, or landscape textures as discriminative cues \citep{vo2016localizing,toker2021coming}. However, image-based methods often face difficulties when visual scenes are visually repetitive, temporally dynamic, or degraded by occlusion, weather, or disaster-related damage, which can obscure geolocation-specific features and limit geolocalization accuracy.

Recently, increasing attention has been devoted to multimodal geolocalization, which integrates text and image information to overcome the limitations of unimodal approaches. In this paradigm, textual cues such as toponyms, directions, or contextual descriptions are combined with visual features extracted from images to provide complementary signals for geolocalization \citep{tahmasebzadeh2024multimodal,yin2025llm,ye2025cross}. Multimodal fusion methods range from simple late-stage integration to advanced cross-modal attention mechanisms that align semantic representations across modalities \citep{radford2021learning,li2023blip}. This combined framework is particularly valuable in disaster scenarios, where text from eyewitnesses or emergency reports may be incomplete or ambiguous, while images can provide concrete yet visually challenging evidence. By jointly leveraging textual semantics and visual appearance, multimodal geolocalization improves robustness in resolving toponym ambiguity, narrowing candidate search spaces, and enhancing fine-grained geolocalization accuracy. Image-based approaches also remain effective, especially when leveraging cross-view visual consistency, providing a practical alternative when textual information is limited or unavailable \citep{li2025cross,yin2025triple, hong2023cross}.

\subsection{Toponym Disambiguation Methods}
\textcolor{black}{Toponym disambiguation addresses the problem of mapping ambiguous toponyms to their correct geographic referents. Traditional methods have largely relied on gazetteers, rule-based heuristics, and spatial constraints, often combining string matching with hierarchical filters such as country, region, or city \citep{middleton2018location,fize2021deep}. While effective in well-structured text, these methods often struggle with user-generated content, where misspellings, abbreviations, colloquial expressions, and incomplete descriptions are common. Probabilistic models and statistical learning techniques were later introduced to incorporate contextual cues such as co-occurring toponyms, nearby entities, or surrounding text \citep{buscaldi2008conceptual,roberts2010toponym}. However, their performance remained limited by data sparsity, weak semantic representations, and lack of adaptability across domains.}

\textcolor{black}{The emergence of neural language models has significantly advanced toponym disambiguation research. Transformer-based models such as BERT \citep{devlin2019bert} and GPT \citep{radford2018improving} capture contextual semantics more effectively than traditional embeddings, improving the recognition of implicit and ambiguous geographic cues. Recent works have enhanced these models with external knowledge sources, such as geographic knowledge graphs, structured gazetteers, or knowledge-enhanced embeddings, to strengthen spatial reasoning and interpretability \citep{buscaldi2008map,gritta2018melbourne}. For example, \citet{delozier2015gazetteer} introduced TopoCluster, a gazetteer-independent method that ranks candidates using geographic word profiles learned from distributional evidence, reducing reliance on exact string matching when gazetteer coverage is sparse or noisy. \citet{hu2024toponym} demonstrates that lightweight open-source MLLMs augmented with geo-knowledge and voting ensembles can deliver competitive resolution across diverse benchmarks. Their pipeline emphasizes efficiency and interpretability, making it suitable for time-sensitive scenarios such as crisis monitoring. In addition, several studies have explored integrating MLLMs with spatial reasoning, user metadata, or temporal information to refine disambiguation results \citep{wang2020neurotpr, sun2023cross, hu2024toponym}.}

\textcolor{black}{Ambiguous or common toponyms, such as street names, chain stores, and public facilities, remain difficult to resolve using text alone. Although existing toponym disambiguation studies have made progress with textual context, gazetteers, and geographic knowledge, they still mainly rely on semantic information and rarely exploit visual evidence from remote sensing or street-view imagery. This limits their ability to distinguish between multiple plausible geolocations associated with the same toponym, especially in complex scenarios such as disasters, where contextual information is often noisy or incomplete. To address this limitation, we propose DisasterTD, a novel framework that integrates MLLM-based semantic candidate generation with cross-view visual verification for disaster toponym disambiguation.}

\section{Methods}
\subsection{Overall Workflow}  
\begin{figure*}[htbp]
\centering
\includegraphics[width=\linewidth]{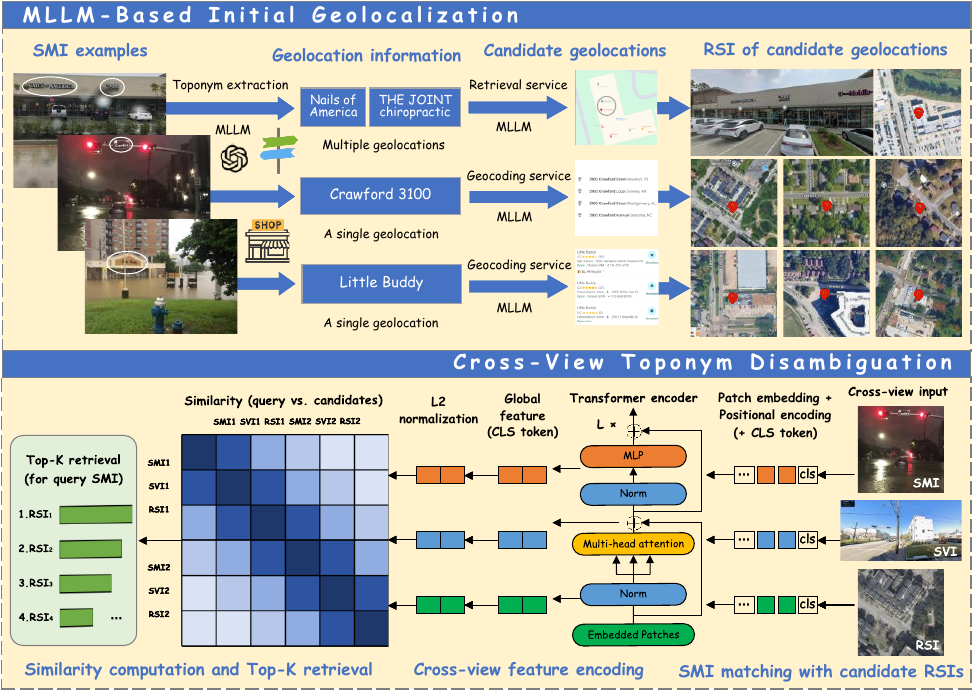}
\vspace{-10pt}
\caption{\textcolor{black}{Overview of DisasterTD, a novel RSI-assisted framework for disaster toponym disambiguation. The pipeline integrates MLLM-based semantic candidate generation with cross-view visual verification, where SMIs are matched with candidate RSIs and SVI serves as an optional bridge.}}
\label{OverallProcess}
\end{figure*}

This study proposes DisasterTD, a novel framework for disaster toponym disambiguation that integrates MLLM-based initial geolocalization with cross-view matching, as shown in Fig. \ref{OverallProcess}. In the first stage, an MLLM (GPT-4o-2024-05-13) is used to extract toponyms from SMI and generate candidate geolocations, which are then mapped to geographic coordinates through map service queries. In the second stage, cross-view matching is applied to refine these candidates. Using panoramic SVI as an intermediate bridge for cross-view matching between ground-level SMI and overhead RSI, mitigating the viewpoint gap and improving the geolocalization accuracy of SMI. For images with a single clear toponym, the corresponding candidate RSIs are compared with the input SMI based on visual similarity to filter out incorrect matches and retain the most plausible geolocation. For images containing vague or multiple toponyms (e.g., common toponyms such as “KFC” or “McDonald’s”), cross-view matching is further employed to identify the most plausible geolocation. By combining these two stages, the proposed framework effectively mitigates errors from toponym ambiguity and improves the reliability and recall of cross-view geolocalization.

\subsection{MLLM-Based Initial Geolocalization}
In disaster scenarios, SMIs often contain useful geographic cues, such as street signs, shop names, building numbers, or short textual messages. However, these cues are often incomplete, occluded, or embedded in complex scenes. To better utilize this information, we use MLLMs to extract text-based toponyms directly from images while considering the surrounding visual context. Given an input image $I$, the MLLM extracts a set of textual tokens or phrases $T = \{t_1, t_2, \ldots, t_m\}$, where each $t_i$ may correspond to a potential geographic cue. Compared with traditional optical character recognition pipelines, MLLMs are more robust to noise and partial text, and can capture semantically meaningful expressions beyond exact string matching, enabling the extraction of both explicit toponyms and implicit geolocation-related information.

The extracted textual set $T$ is then mapped to a candidate toponym set $G = \{G_1, G_2, \ldots, G_n\}$, where each $G_i$ represents a possible geographic entity. These entities differ in specificity: a full address unique point of interest (POI) often corresponds to a single geolocation, while generic names (e.g., chain stores) may correspond to multiple possible geolocations. Accordingly, images are categorized based on the structure of $G$. When $|G| = 1$ and the toponym is distinctive, the mapping is relatively direct. When $|G| > 1$ or the toponyms are ambiguous, spatial relationships among candidates are considered. This can be formulated as selecting a geolocation l that maximizes consistency:

\begin{equation}
l^* = \arg\max_{l} \sum_{i=1}^{n} \mathbb{I}(l \in \mathcal{L}(G_i)),
\end{equation}
where $\mathcal{L}(G_i)$ denotes the set of possible geolocations associated with $G_i$.

Each toponym $G_i$ is linked to external map services to retrieve candidate geolocations. For single, well-defined toponyms, a geocoding function $f_{\text{geo}}: G \rightarrow \mathbb{R}^2$ maps the entity directly to coordinates $(lat, lon)$. For ambiguous or multiple toponyms, a place retrieval function $f_{\text{place}}: G \rightarrow \{l_1, l_2, \ldots, l_k\}$ generates a candidate set of geolocations. To reduce irrelevant results, the search can be restricted to a disaster-affected region $\Omega$, i.e., $l_j \in \Omega$. This stage outputs a candidate geolocation set $L = \{l_1, l_2, \ldots, l_k\}$ for each image, which is further refined in the subsequent cross-view matching stage. This initial geolocalization strategy follows our previous work \citep{yin2025llm}, and more details can be found therein.


\subsection{Cross-View Toponym Disambiguation}
\subsubsection{Toponym Disambiguation Strategies}
We apply different strategies to different types of toponyms during disambiguation. For SMIs containing a single clear toponym, we first use the MLLM recognition results to obtain initial candidate geolocations and retrieve the corresponding RSIs. SVI is introduced as a bridge to reduce the difficulty caused by cross-view differences. Specifically, SMI, SVI, and RSI are encoded into high-dimensional feature vectors using a ViT-based visual foundation model \cite{hong2024spectralgpt}, DINOv2, and cosine similarity is then used to measure matching scores. If a candidate RSI shows high similarity with the input SMI, its geolocation is confirmed as the disambiguation result; otherwise, the process continues with the remaining candidates until the optimal geolocation is identified. For SMIs containing ambiguous or multiple toponyms (e.g., “KFC” or “McDonald’s”), the search range is relaxed during the initial retrieval stage to generate multiple candidate geolocations. The corresponding RSI for these candidates are then obtained, and with SVI as a bridge, each candidate RSI is matched with the SMI through DINOv2-based feature encoding and similarity computation. The geolocation associated with the RSI that is most similar to the input SMI is finally selected as the disambiguation result.

\subsubsection{DINOv2-Based Cross-View Matching}


In cross-view geolocalization, there exists a significant domain gap between SMI and RSI, and direct matching often fails to achieve satisfactory results. To address this issue and improve the accuracy of toponym disambiguation, we extend our previously proposed cross-view geolocalization method \citep{yin2025triple}. This method was originally applied to SMI-based geolocalization in disaster scenarios, and in this study it is introduced for the first time into the task of toponym disambiguation. We introduce SVI as an optional intermediate bridge and construct a three-view joint optimization framework of SMI$\leftrightarrow$SVI$\leftrightarrow$RSI. \textcolor{black}{As shown in Fig. \ref{OverallProcess}, we use the ViT-based DINOv2 as the backbone for cross-view feature extraction. Each input image, including the query SMI, optional SVI, and candidate RSIs, is divided into image patches, combined with positional encoding and a CLS token, and processed by Transformer encoder blocks to obtain CLS-based global features. After L2 normalization, cosine similarity is computed between the query SMI and each candidate RSI, and the candidates are ranked for Top-K retrieval.} By jointly constraining local semantics and global spatial consistency through a multi-objective loss, the method enhances SMI$\leftrightarrow$RSI matching and provides robust support for toponym disambiguation in complex scenarios.

We use a DINOv2 model based on the ViT architecture \citep{dosovitskiy2020image} for cross-view feature encoding. Given an input image $I$, the model divides it into patches, embeds them into a sequence, and encodes them through multi-layer self-attention to obtain a high-dimensional representation:

\begin{equation}
z = \mathrm{DINOv2}(I;\theta), \quad z \in \mathbb{R}^d
\end{equation}
where $\theta$ denotes the trainable parameters and $d$ is the feature dimension.

To achieve cross-view alignment, features from the three views are jointly optimized with the following objective:
\begin{equation}
\mathcal{L}_{total} = \lambda_1 \mathcal{L}(z_m, z_s) + \lambda_2 \mathcal{L}(z_s, z_r) + \lambda_3 \mathcal{L}(z_m, z_r)
\end{equation}

where $I_m, I_s, I_r$ are the candidate SMI, SVI, and RSI, and $z_m, z_s, z_r$ are their corresponding feature vectors. $\mathcal{L}(\cdot)$ is the InfoNCE contrastive loss, $\lambda_1, \lambda_2, \lambda_3$ are weighting coefficients, and $\mathcal{L}_{total}$ is the overall optimization objective.

During training, we initialize the model with DINOv2 self-supervised pretraining weights \citep{dosovitskiy2020image, oquab2023dinov2} and use the model trained on the MultiIan dataset collected in disaster scenarios to improve generalization on large-scale unlabeled images. DINOv2 enables the model to learn semantically rich and domain-invariant features by enforcing consistency between representations of different augmented views of the same image without requiring manual labels. This is achieved by minimizing a cross-view consistency loss between the teacher and student outputs \citep{oquab2023dinov2}, which can be formulated as:

\begin{equation}
\mathcal{L}_{DINO} = - \sum_{i=1}^{N} P^{(t)}(x_i) \cdot \log P^{(s)}(x_i)
\end{equation}

Here, $\mathcal{L}_{DINO}$ denotes the cross-entropy loss, where $P^{(t)}(x_i)$ and $P^{(s)}(x_i)$ are the teacher and student output distributions for the $i$-th feature, and $N$ is the number of feature dimensions. This objective encourages the student network to align its outputs with those of the teacher across varying augmentations, resulting in stable and semantically consistent representations.

Moreover, to complement the global consistency enforced by the class token, DINOv2 incorporates a patch-level objective inspired by iBOT \citep{caron2021emerging}. The loss is defined as:

\begin{equation}
\mathcal{L}_{\text{iBOT}} = - \sum_{i=1}^{M} P^{(t)}_i \cdot \log P^{(s)}_i
\end{equation}

where $P^{(t)}_i$ and $P^{(s)}_i$ represent the teacher and student distributions over the masked patch tokens at position $i$, and $M$ is the number of masked patches. This loss promotes local feature alignment between corresponding regions, improving the model’s capacity to learn fine-grained representations. In DisasterTD, the same encoder architecture is used for both SVI and RSI to ensure consistent feature extraction and facilitates end-to-end joint training, ultimately improving geolocalization performance and robustness in complex scenarios.

To increase sensitivity to local details, the oveall training objective combines both image-level and patch-level constraints, and incorporates KoLeo regularization to encourage uniform feature distribution:

\begin{equation}
\mathcal{L}_{KoLeo} = -\frac{1}{N}\sum_{i=1}^{N} \log \left( \min_{j \neq i} \| z_i - z_j \| \right)
\end{equation}

where $N$ is the number of samples, $z_i$ denotes the feature vector of the $i$-th sample, $z_j$ denotes other feature vectors, and $| z_i - z_j |$ is the Euclidean distance between features.


\subsection{Baselines and Accuracy Evaluation}
To comprehensively evaluate the effectiveness of DisasterTD, we designed two types of baseline experiments. 1) MLLM-based geolocalization only: GPT-4o is directly used to extract toponyms from SMI, and geographic coordinates are obtained through Google Maps Geocoding or Places Search services, without any cross-view validation or filtering. This baseline reflects the role of MLLM in toponym recognition and initial geolocalization. 2) Cross-view matching only: geolocation is inferred by directly matching SMI with RSI, and we further compare the effect of introducing SVI as an intermediate bridge. This baseline relies entirely on cross-view visual consistency, without MLLM-based toponym recognition or candidate generation. Within this framework, we compare two setting in the cross-view disambiguation stage: 1) direct matching between SMI and RSI; 2) joint cross-view matching among SMI, SVI, and RSI, where SVI serves as an intermediate bridge. \textcolor{black}{To further evaluate whether the performance gains come from the proposed semantic candidate-generation and visual-verification framework rather than from a specific visual encoder, we also include several representative cross-view geolocalization models, including ConvNeXt \cite{liu2022convnet}, SAIG-D \cite{zhu2023simple}, TransGeo \cite{zhu2022transgeo}, and Sample4Geo \cite{deuser2023sample4geo}. These models are evaluated in two ways: replacing only the second-stage visual matching module of DisasterTD while keeping the MLLM-based candidate generation stage unchanged, and directly applying them to SMI$\leftrightarrow$RSI geolocalization without the proposed candidate-generation and disambiguation framework.}


\begin{table}[t]
\centering
\setlength{\tabcolsep}{8pt}
\caption{Geolocalization distance thresholds and corresponding spatial levels.}
\label{tab:distance_levels}
\begin{tabular}{cccc}
\toprule
Distance & Spatial Level & Description & Granularity\\
\midrule
50 m   & POI-level           & Building or POI & Fine-grained\\
100 m  & Street-level        & Street segment & Fine-grained \\
250 m  & Block-level         & Urban block & Medium\\
500 m  & Neighborhood-level  & Neighborhood  & Medium\\
1000 m & District-level      & Urban area   & Coarse\\
\bottomrule
\end{tabular}
\end{table}

We use Geoloc ACC@k (k = 1000, 500, 250, 100, 50 m) as the core metric to evaluate geolocalization performance at multiple spatial scales, as defined in Equation \ref{GeolocACC_accuracy}. This metric measures the proportion of samples whose predicted geolocation falls within a radius of $k$ meters from the ground truth. \textcolor{black}{As shown in Table \ref{tab:distance_levels}, different distance error thresholds represent different spatial levels in disaster response, ranging from coarse district- or neighborhood-level awareness to fine-grained street- or POI-level geolocalization.} Since cross-view retrieval methods are often evaluated with Recall@k, we unify them under the Geoloc ACC metric: the predicted geolocation (Top-1 or Top-k) is selected from the retrieval results, its geodesic distance to the ground truth is computed, and compared with the threshold. This yields Geoloc ACC@1000 m, Geoloc ACC@500 m, Geoloc ACC@250 m, Geoloc ACC@100 m, and Geoloc ACC@50 m. In addition, we report Top-1 mean and median spatial errors (in meters) to complement the threshold-based metrics, providing a continuous measure of geolocalization accuracy and robustness.

\begin{equation}
Geoloc\ ACC@d\,\text{m} = \frac{\left| \{ D_{\text{pre}} \mid \text{dist}(D_{\text{pre}}, D_i) < d \} \right|}{\sum_{j=1}^{n} |D_j|}
\label{GeolocACC_accuracy}
\end{equation}

\noindent In this metric, a prediction \( D_{\text{pre}} \) is regarded as correct if its geodesic distance from the ground truth \( D_i \) is smaller than a predefined threshold \( d \) m. The numerator counts the number of samples that satisfy this criterion, while the denominator corresponds to the total number of evaluated samples. The resulting ratio reflects the proportion of accurate predictions under the specified distance constraint.

\section{Results}
\subsection{Data Preparation}

\begin{figure}[b]
\centering
\includegraphics[width=\linewidth]{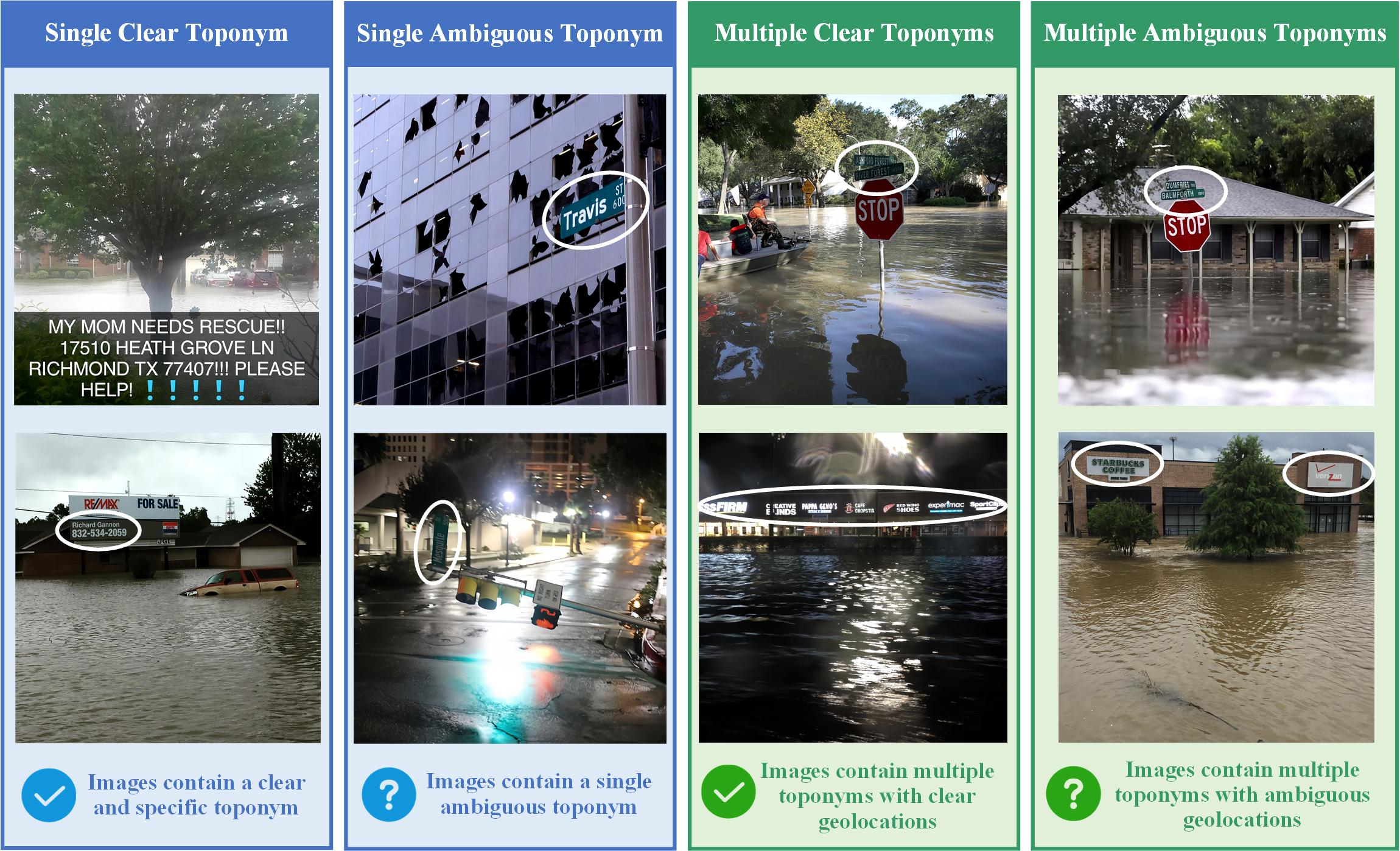}
\vspace{-10pt}
\caption{Examples of different SMI types, including images with single or multiple toponyms referring to clear or ambiguous geolocations.}
\label{DataExamples}
\end{figure}


The disaster-related SMI used in this study are derived from the Hurricane Harvey Twitter dataset \citep{Phillips2024Hurricane}. This dataset contains a large number of publicly shared SMI during Hurricane Harvey in the United States in 2017, covering diverse scene content and geographic information. We use 1,000 SMI with clear geographic cues that were randomly selected in our previous work \citep{yin2025llm}. The scale is comparable to datasets used in existing research \citep{hu2023geo,yin2025llm} and is sufficient for this study. The SMIs are categorized based on the amount and type of geographic information, as shown in Fig. \ref{DataExamples}. Images with a single clear toponym (272 samples as examples) include SOS messages, phone numbers, street signs, or a single POI, among which 134 are SOS or phone number cases and 138 contain only a single street sign or POI. Images with a single ambiguous toponym (275 samples) mainly consist of the remaining single street sign or single POI cases, where the toponym often corresponds to multiple candidate geolocations with higher ambiguity. Images with multiple clear toponyms (302 samples) contain the co-occurrence of several street signs or POIs, where the toponyms are relatively unambiguous. Finally, images with multiple ambiguous toponyms (151 samples) involve several ambiguous toponyms appearing simultaneously, representing complex cases that are difficult to resolve directly.

\begin{figure*}
\centering
\includegraphics[width=\linewidth]{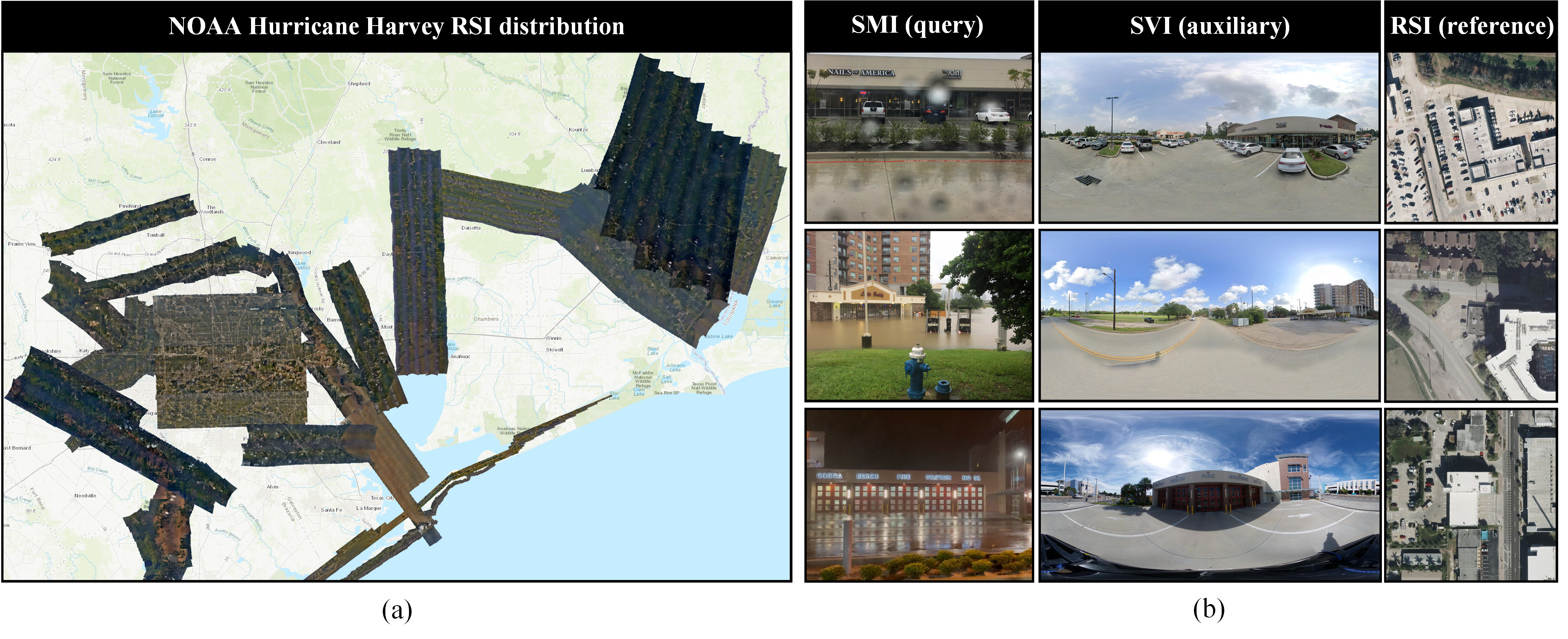}
\vspace{-20pt}
\caption{Data examples. Spatial distribution of NOAA Hurricane Harvey RSI data (a), and examples of cross-view toponym disambiguation samples, including query SMI, candidate RSI, and auxiliary SVI (b).}
\label{Data}
\end{figure*}

To conduct cross-view toponym disambiguation experiments, we retrieved the candidate RSI and SVI corresponding to each SMI, thereby constructing a multi-view aligned dataset. Data examples are shown in Fig. \ref{Data}. RSI from the National Oceanic and Atmospheric Administration (NOAA) provide information on overall spatial structure and environmental patterns, which help capture macro-level geographic features. SVI serve as a bridge between SMI and RSI, playing a key role in modeling fine-grained local semantics and object appearance, and effectively reducing the difficulties caused by cross-view differences. By integrating SMI, SVI, and RSI, we constructed a cross-view image dataset for Hurricane Harvey, providing a solid foundation for subsequent cross-view matching and toponym disambiguation.

\vspace{-10pt}

\subsection{MLLM-Based Initial Geolocalization Performance}

The MLLM-based toponym recognition results are shown in Table \ref{tab:InitialGeolocalization}. The method achieves strong performance across different image types, with an overall accuracy of 88.35\%. Building on this, the MLLM-based initial geolocalization method parses toponyms in SMI into candidate geolocations and evaluates accuracy at different spatial thresholds. Table \ref{tab:InitialGeolocalization} also reports the geolocalization results under thresholds of 1000 m, 500 m, 250 m, 100 m, and 50 m, with the mean and median distance errors. As expected, the accuracy gradually decreases as the threshold becomes stricter, since tighter spatial constraints reduce the tolerance of candidate sets and may exclude some true geolocations. The effectiveness of this method has also been demonstrated to outperform baselines that rely solely on MLLMs without map services, those that rely only on map services without MLLMs, as well as Geoapify and Nominatim \cite{yin2025llm}.

Across all thresholds, SMI with a single clear toponym achieve the best performance, reaching 60.45\% at 1000 m and remaining relatively robust at finer scales (49.02\% at 250 m and 41.50\% at 50 m). This indicates that in cases with strong directional cues, such as SOS messages or phone numbers, MLLM-based initial geolocalization can effectively generate candidates close to the ground truth. The accuracy for multiple clear toponyms is 59.39\%, comparable to single clear toponyms, suggesting that when multiple unambiguous toponyms co-occur, the candidate range can still be well constrained. In contrast, single ambiguous toponyms perform the worst, with an accuracy of only 29.69\%, dropping further to 20.21\% at 250 m and 10.01\% at 50 m. Such vague toponyms (e.g., chain restaurants or common venue names) often correspond to a large number of widely distributed candidates, making fine-grained geolocalization highly challenging. Multiple ambiguous toponyms achieve slightly higher accuracy of 31.60\%, showing that although ambiguity exists, the combination of several vague toponyms can provide weak constraints and lead to a modest improvement.

At stricter thresholds, the decline in accuracy becomes more pronounced. For single clear toponyms, accuracy drops from 60.45\% to 52.10\% (500 m), 49.02\% (250 m), 45.37\% (100 m), and 41.50\% (50 m). For single ambiguous toponyms, accuracy decreases sharply from 29.69\% to 22.48\%, 20.21\%, 11.25\%, and 10.01\%. Even in single clear toponym cases, not all results can be resolved precisely, as errors may still occur during text recognition and geocoding. In such cases, the predicted candidates may be close to the true geolocation but still fall outside stricter thresholds. For multi-toponym cases, the initial accuracy is usually higher than that of single ambiguous toponyms, but as thresholds tighten, the complexity introduced by multiple candidates also causes accuracy to drop steadily. This suggests that while co-occurrence of multiple toponyms can provide spatial constraints, it cannot fully eliminate error accumulation and may even complicate the geolocalization problem. Overall accuracy decreases from 47.31\% at 1000 m to 39.17\% (500 m), 36.51\% (250 m), 31.19\% (100 m), and 27.94\% (50 m), highlighting the limitations of relying solely on MLLM-based initial geolocalization under strict spatial constraints and underscoring the necessity of cross-view matching for further disambiguation.

\begin{figure*}
\centering
\includegraphics[width=\linewidth]{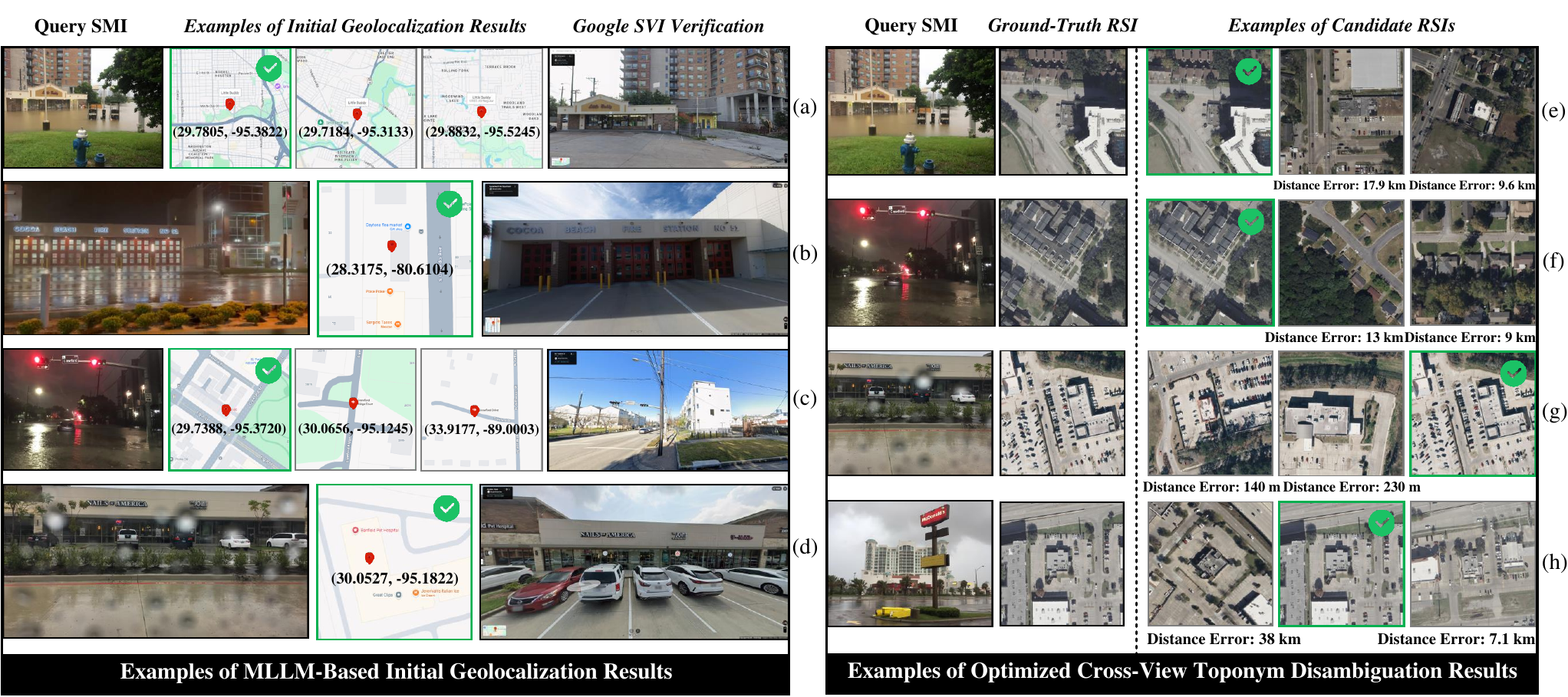}
\vspace{-10pt}
\caption{Examples of MLLM-based initial geolocalization results and optimized cross-view toponym disambiguation results. SMIs in (a), (c), (e), and (f) contain single ambiguous toponyms; image (b) contains a single clear toponym; images in (d) and (g) contain multiple clear toponyms; and image (h) contains multiple ambiguous toponyms. Candidate geolocations with green borders are correct, while those with gray borders are incorrect.}
\label{ResultsExamples}
\end{figure*}

As shown in Fig. \ref{ResultsExamples}, SMI containing only ambiguous toponyms often produce retrieval results with spatially dispersed candidates. This observation confirms that MLLM-based initial geolocalization is more suitable as a candidate generation step. Its core value lies in extracting a possible spatial range from complex semantic information. Although ambiguity cannot be completely resolved, the true geolocation is usually included within the candidate set, providing essential input and data support for subsequent cross-view matching and toponym disambiguation. Even when the retrieved candidates are widely distributed, this method still significantly reduce the search space compared to exhaustive geographic retrieval.

\begin{table*}
\centering
\caption{\textcolor{black}{MLLM-based toponym recognition and geolocalization accuracies across image categories and distance thresholds. The types include images with single-clear, single-ambiguous, multiple-clear, or multiple-ambiguous toponyms, and overall performance across categories. The optimal value for each metric is shown in bold.}}
\label{tab:InitialGeolocalization}
\footnotesize
\begin{tabular}{lcccccccc}
\toprule
\diagbox{Type}{Metric} & RecogACC (\%) &  \makecell{Geoloc ACC\\@1000 m (\%)} &  \makecell{Geoloc ACC\\@500 m (\%)} &  \makecell{Geoloc ACC\\@250 m (\%)} & \makecell{Geoloc ACC\\@100 m (\%)} & \makecell{Geoloc ACC\\@50 m (\%)} & \makecell{Mean \\Error (km)} & \makecell{Median \\Error (km)}\\
\midrule
Single-Clear       & 88.58 & \textbf{60.45} & \textbf{52.10} & \textbf{49.02} & \textbf{45.37} & \textbf{41.50} & 14.63 & \textbf{0.88}\\
Single-Ambiguous   & \textbf{92.17} & 29.69 & 22.48 & 20.21 & 11.25  & 10.01 & 28.29 & 1.85\\
\vspace{1pt}
Overall-Single & 90.38 & 44.99 & 37.21 & 34.54 & 28.22 & 25.67 & 21.50 & 1.37\\

Multiple-Clear     & 86.29 & 59.39 & 50.07 & 47.88& 45.21  & 40.80 & \textbf{13.25} & 0.91\\
Multiple-Ambiguous & 85.10 & 31.60 & 24.45 & 20.93& 13.90  & 10.44& 22.88 & 1.46\\
\vspace{1pt}
Overall-Multiple & 85.89 & 50.13 & 41.53 & 38.90 & 34.77 & 30.68 & 16.46 & 1.09\\

Overall Performance  &  88.35 & 47.31 & 39.17 & 36.51 & 31.19  & 27.94 & 19.22 & 1.24 \\
\bottomrule
\end{tabular}
\end{table*}

\subsection{DisasterTD Performance with Cross-View Disambiguation}
\subsubsection{Single-Toponym Disambiguation Performance}

For images containing a single toponym, MLLM-based initial geolocalization can usually provide a candidate set that contains the true geolocation, but the direct geolocalization accuracy at fine scales is limited. Fig. \ref{ResultsExamples} (e)–(h) illustrates representative examples, where each query SMI is associated with several candidate RSIs. The green box indicates the correctly matched RSI, while the others correspond to incorrect candidates with varying spatial deviations. Fig. \ref{ResultsExamples} (e) shows an example containing the ambiguous toponym “Little Buddy”, where the candidate RSIs are distributed across different geolocations, with distance errors of 17.9 km and 9.6 km for incorrect matches. Fig. \ref{ResultsExamples} (f) presents another ambiguous case with the toponym “Crawford 3100”, where incorrect candidates exhibit distance errors of 13 km and 9 km. These examples highlight that ambiguous toponyms often correspond to geographically dispersed candidates, making accurate geolocalization difficult without additional cross-view constraints.



After introducing cross-view matching, the geolocalization accuracy of SMI improves significantly at all spatial thresholds. As shown in Table \ref{tab:DisambiguationSingle}, the accuracy of single clear toponyms increases from 60.45\% to 78.52\% at the 1000 m threshold, and to 70.35\%, 66.26\%, 61.70\%, and 55.97\% at the 500 m, 250 m, 100 m, and 50 m thresholds, respectively. This demonstrates that while MLLM-based initial geolocalization already generates reasonable candidates near the ground truth, cross-view matching effectively removes incorrect options. The improvement is even more pronounced for single ambiguous toponyms: accuracy rises sharply from 29.69\% to 65.23\% at 1000 m, and to 55.12\%, 49.36\%, 42.38\%, and 36.08\% at 500 m, 250 m, 100 m, and 50 m. This shows that cross-view matching plays a crucial role when handling images with vague toponyms (e.g., chain stores or common venue names), as it can filter a large number of widely distributed candidates and significantly reduce errors caused by ambiguity.

Further analysis shows that the impact of cross-view matching varies across categories. For single clear toponyms, the improvement is moderate and mainly serves to validate and refine initial candidates. For single ambiguous toponyms, where the initial accuracy is low, the cross-view mechanism brings much larger gains, effectively reducing errors caused by toponym ambiguity. Overall performance in single-toponym scenarios reaches 71.84\% at 1000 m and remains at 62.69\%, 57.71\%, 51.99\%, and 45.97\% under stricter thresholds, showing consistent improvements of over 20\% compared to the initial results. The mean error is reduced to 11.06 km, with a median error of 0.70 km, indicating that cross-view matching improves accuracy while reducing large spatial deviations.

\begin{table*}
\centering
\caption{\textcolor{black}{Geolocalization accuracies of DisasterTD across image categories and distance thresholds. The types include images with single-clear, single-ambiguous, multiple-clear, or multiple-ambiguous toponyms, and overall performance across categories. The optimal value for each metric is shown in bold.}}
\label{tab:DisambiguationSingle}
\footnotesize
\begin{tabular}{lccccccc}
\toprule
\diagbox{Type}{Metric} & \makecell{Geoloc ACC\\@1000 m (\%)} &  \makecell{Geoloc ACC\\@500 m (\%)} &  \makecell{Geoloc ACC\\@250 m (\%)} & \makecell{Geoloc ACC\\@100 m (\%)} & \makecell{Geoloc ACC\\@50 m (\%)} & Mean Error (km) & Median Error (km)\\
\midrule
Single-Clear       & \textbf{78.52} & \textbf{70.35} & \textbf{66.26} & \textbf{61.70} & \textbf{55.97} & \textbf{10.34} & \textbf{0.60} \\
Single-Ambiguous   & 65.23 & 55.12 & 49.36 & 42.38   & 36.08 & 11.77 & 0.80\\
\vspace{1pt}
Overall-Single   & 71.84 & 62.69 & 57.71& 51.99    & 45.97 & 11.06 & 0.70\\

Multiple-Clear    & 75.80 & 67.91 & 64.70 & 58.99   & 55.51 & 10.83 & 0.72\\
Multiple-Ambiguous  & 62.45 & 50.04  &  45.41& 38.66   & 33.80 & 13.29 & 0.55\\
\vspace{1pt}
Overall-Multiple  & 71.35 & 61.95 & 58.27 & 52.21  & 48.27 & 11.65 & 0.66\\
Overall Performance  & 71.62 & 62.36 & 57.99 & 52.09 & 47.01 & 11.33 & 0.68\\
\bottomrule
\end{tabular}
\end{table*}

\subsubsection{Multi-Toponym Disambiguation Performance}
In multi-toponym scenarios, cross-view matching also leads to significant improvements in geolocalization accuracy. Fig. \ref{ResultsExamples} (g) shows a case containing the toponyms “NAILS OF AMERICA” and “THE JOINT chiropractic”, where the incorrect candidate RSIs are also close to the ground truth (e.g., 140 m and 230 m), indicating that multiple relatively distinctive toponyms can effectively constrain the search space. Fig. \ref{ResultsExamples} (h) includes the toponyms “McDonald's”, “Chili's”, and “SUBWAY”, where incorrect candidates exhibit large spatial deviations (e.g., 38 km), while the correct geolocation is successfully identified through cross-view matching. These examples show that although MLLM-based initial geolocalization can include the true geolocation, the candidates may be widely dispersed, and cross-view matching is essential for filtering out inconsistent results. As reported in Table \ref{tab:DisambiguationSingle}, for multiple clear toponyms, accuracy increases from 59.39\% to 75.80\% at 1000 m, from 50.07\% to 67.91\% at 500 m, and from 45.21\% to 64.70\% at 250 m. The improvements remain consistent at finer scales (58.99\% at 100 m and 55.51\% at 50 m), showing that cross-view matching effectively leverages spatial consistency among multiple toponyms to maintain robust performance under stricter thresholds.


SMI with multiple ambiguous toponyms present a more challenging case. Such toponyms often correspond to a large number of widely distributed candidates, making MLLM-based retrieval alone insufficient. With cross-view matching, spatial and visual consistency across candidates can be evaluated, leading to substantial performance gains. Accuracy increases from 31.60\% to 62.45\% at 1000 m, from 24.45\% to 50.04\% at 500 m, and from 13.90\% to 38.66\% at 100 m, with consistent improvements at 250 m (45.41\%) and 50 m (33.80\%). These results show that cross-view matching significantly reduces errors caused by ambiguity and improves the reliability of geolocalization.

\begin{figure*}[htbp]
\centering
\includegraphics[width=\linewidth]{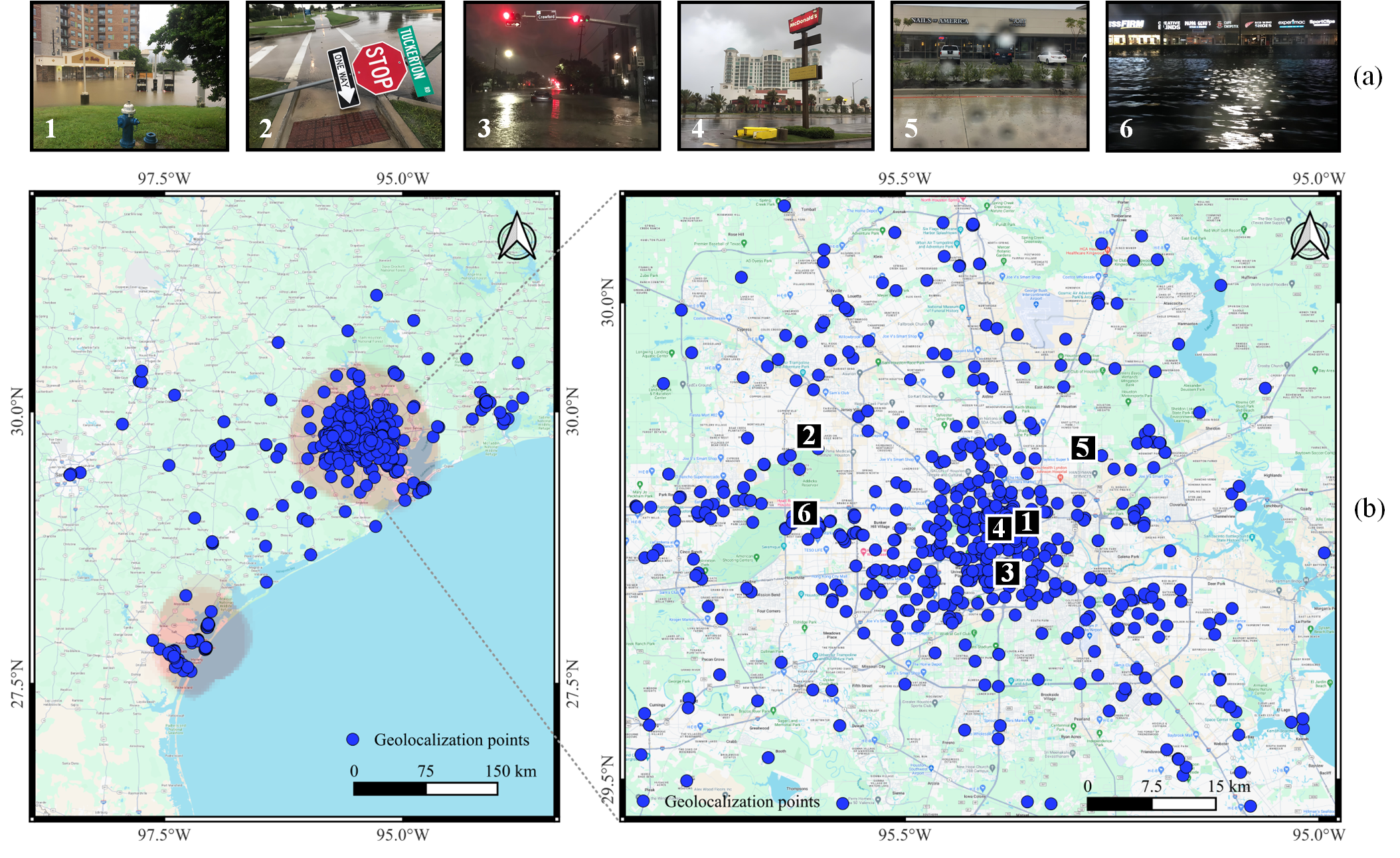}
\vspace{-15pt}
\caption{Visualization of disaster geolocalization results within a 1000 m error range. Example SMIs in (a) correspond to the geolocalization points shown in the zoomed-in map view in (b).}
\label{Visualization}
\end{figure*}

The average accuracy in multi-toponym scenarios reaches 71.35\%, 61.95\%, 58.27\%, 52.21\%, and 48.27\% at 1000 m, 500 m, 250 m, 100 m, and 50 m, respectively, with improvements comparable to those in single-toponym scenarios. The mean error is reduced to 11.65 km, with a median error of 0.66 km, further indicating improved spatial precision. In multi-toponym cases, cross-view matching plays a dual role: for multiple clear toponyms, it mainly refines and validates candidates, while for multiple ambiguous toponyms, it becomes the key mechanism for identifying the correct geolocation from a large and dispersed candidate set. Fig. \ref{Visualization} provides a visualization of correctly localized points at the 1000 m threshold, further demonstrating the effectiveness of the proposed DisasterTD.

\textcolor{black}{The gap between the mean and median errors suggests a long-tailed error distribution. Most predictions are close to the ground truth, as reflected by the low median error of 0.68 km, while a few severe failures increase the mean error to 11.33 km. These large-error cases mainly occur when candidate geolocations are highly dispersed, the correct geolocation is missing or poorly represented, or visually similar buildings, road segments, and surrounding layouts cause incorrect cross-view matching. This suggests that DisasterTD is effective for most samples, but its reliability can still be affected by candidate completeness and visual ambiguity in difficult cases.}

\subsection{Baseline Performance Comparison}

The MLLM-based initial geolocalization method described earlier serves as the baseline for this study. The MLLM-only approach achieves accuracies of 47.31\%, 39.17\%, 36.51\%, 31.19\%, and 27.94\% at 1000 m, 500 m, 250 m, 100 m, and 50 m, respectively, which are clearly lower than those of cross-view methods. This result indicates that relying solely on semantic candidate generation is insufficient for reliable fine-grained geolocalization. To further examine the role of visual matching, we introduce two direct cross-view geolocalization baselines, namely SMI$\leftrightarrow$RSI and SMI$\leftrightarrow$SVI$\leftrightarrow$RSI. The SMI$\leftrightarrow$RSI method achieves 56.84\%, 45.17\%, 40.05\%, 33.92\%, and 28.99\%, consistently outperforming the MLLM-only baseline across all thresholds. When SVI is introduced as an intermediate representation, the performance further improves to 63.26\%, 51.38\%, 46.22\%, 40.57\%, and 35.24\%, suggesting that SVI helps reduce cross-view discrepancies and provides more stable visual correspondence. Despite these improvements, both direct cross-view methods still show noticeable performance degradation as the distance threshold decreases, indicating that image-only matching remains sensitive to viewpoint differences and scene complexity.

\begin{figure*}[htbp]
\centering
\includegraphics[width=\linewidth]{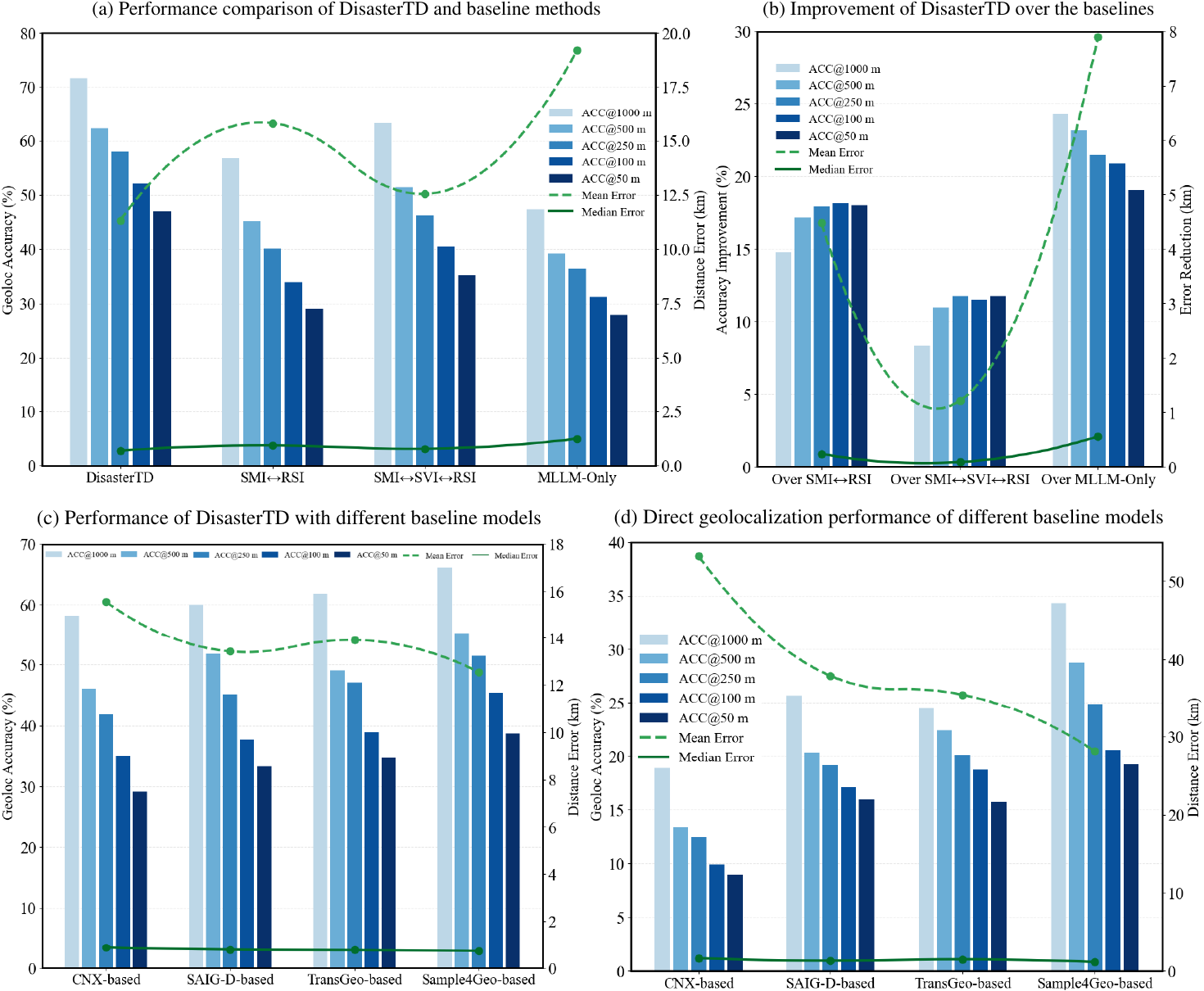}
\vspace{-10pt}
\caption{\textcolor{black}{Geolocalization performance comparison. (a) Performance of DisasterTD and baseline methods. (b) Improvement of DisasterTD over the baselines. (c) Performance of DisasterTD with different cross-view geolocalization models as the visual matching module. (d) Direct SMI-to-RSI geolocalization performance of these baseline models without candidate generation and disambiguation.}}
\label{Baseline}
\end{figure*}

As shown in Fig. \ref{Baseline} (a), the proposed DisasterTD achieves 71.62\%, 62.36\%, 57.99\%, 52.09\%, and 47.01\% across the same thresholds, maintaining clear advantages over all baselines. Compared with direct cross-view methods, the improvement is consistent rather than concentrated at a specific scale, suggesting that the integration strategy contributes across the entire geolocalization range. The advantage becomes more apparent at finer spatial resolutions, where competing methods show a sharper decline. In particular, the accuracy remains above 50\% at 100 m, while both SMI$\leftrightarrow$RSI and SMI$\leftrightarrow$SVI$\leftrightarrow$RSI fall below this level. This behavior indicates that the proposed method is less affected by local ambiguities and can better preserve spatial precision when narrowing down candidate geolocations. The trend is also reflected in Fig. \ref{Baseline} (b), where the improvement margins remain relatively stable across thresholds, instead of fluctuating significantly with distance.

\textcolor{black}{As shown in Fig. \ref{Baseline} (c) and (d), we further evaluate several representative cross-view geolocalization models, including ConvNeXt, SAIG-D, TransGeo, and Sample4Geo. Fig. \ref{Baseline} (c) reports the results when only the second-stage visual matching module of DisasterTD is replaced by these models, while the MLLM-based candidate generation stage remains unchanged. Fig. \ref{Baseline} (d) shows the results when these models are directly applied to SMI$\leftrightarrow$RSI geolocalization without the proposed candidate-generation and disambiguation framework. Across all four models, incorporating them into DisasterTD leads to much higher accuracy than direct SMI$\leftrightarrow$RSI geolocalization. For example, Sample4Geo-based DisasterTD achieves 66.08\%, 55.23\%, 51.50\%, 45.37\%, and 38.68\% from 1000 m to 50 m, whereas direct SMI$\leftrightarrow$RSI Sample4Geo reaches only 34.33\%, 28.82\%, 24.91\%, 20.50\%, and 19.25\%. Similar trends are observed for ConvNeXt, SAIG-D, and TransGeo. These results show that the gains of DisasterTD mainly come from combining MLLM-based semantic candidate generation with cross-view visual verification, rather than from a specific visual encoder.}

The proposed DisasterTD combines candidate generation and filtering in a more structured manner, allowing the system to first narrow down the search space and then perform targeted verification. This design reduces the impact of both semantic ambiguity and visual mismatch, leading to more stable performance across different spatial scales. Compared with direct cross-view matching, which relies entirely on visual similarity, the additional semantic constraint helps eliminate implausible candidates at an earlier stage, making the subsequent matching process more focused. As a result, even under strict geolocalization constraints, the method maintains relatively high accuracy and avoids the rapid degradation observed in baseline approaches. These results suggest that image-only cross-view matching is insufficient to fully resolve toponym ambiguity, and that effective fine-grained geolocalization in disaster scenarios requires the joint use of semantic candidate generation and cross-view disambiguation.

\section{Discussion}

This study presents DisasterTD, a framework that integrates multimodal LLM-based semantic reasoning with cross-view geolocalization for disaster toponym disambiguation. The proposed two-stage framework first generates candidate geolocations using MLLMs and then refines these candidates through cross-view visual matching. By explicitly combining semantic cues with spatial and visual evidence, DisasterTD reduces candidate ambiguity and improves geolocalization accuracy across multiple distance thresholds. Compared with text-only or image-only approaches, the framework provides a more balanced and robust solution: semantic reasoning narrows the search space, while cross-view matching further enforces spatial consistency and visual correspondence. This synergy is particularly important in disaster scenarios, where information is often incomplete, heterogeneous, and time-sensitive. The use of a ViT-based visual foundation model further enhances feature alignment across heterogeneous views, enabling more reliable matching between SMI, SVI, and RSI. DisasterTD provides a practical and scalable solution for fine-grained geolocalization under severe toponym ambiguity, addressing an important challenge in disaster-related GeoAI applications.

Despite these advantages, several limitations remain. The current evaluation is mainly conducted on the Hurricane Harvey dataset, which may limit the generalizability of DisasterTD to other disaster types, geographic regions, or linguistic contexts. Disaster-related imagery and textual expressions vary considerably across events, and the model may be sensitive to such domain shifts and regional differences. The framework also relies on external map services for candidate generation and SVI retrieval, which may introduce biases due to uneven global coverage, differences in data quality, or missing information in less-mapped regions. In practice, these limitations may reduce the completeness of candidate sets and propagate errors to the subsequent cross-view matching stage.

\textcolor{black}{Temporal inconsistency is another important challenge. SMI collected during disasters often capture rapidly changing scenes, while the corresponding RSI or SVI may be acquired at different times, leading to discrepancies in visual appearance and environmental conditions. Such temporal gaps can weaken cross-view matching performance, especially in heavily damaged or evolving environments. From a computational perspective, the SMI$\leftrightarrow$SVI$\leftrightarrow$RSI retrieval process can become costly when the candidate pool is large, requiring multiple rounds of feature extraction and similarity computation. Since this study does not include a systematic runtime, latency, or throughput benchmark, the current evaluation focuses mainly on geolocalization accuracy rather than operational efficiency. This may limit scalability and delay response time in real-world emergency scenarios where efficiency is critical. These observations suggest that further improvements are still needed in data coverage, temporal alignment, retrieval efficiency, and model robustness to support large-scale practical deployment.}


Future work can further extend DisasterTD in several directions. Evaluating DisasterTD on a wider range of disaster events and multilingual datasets would help better understand its behavior under different conditions and reduce potential bias toward a single benchmark. Incorporating temporal information, such as event timelines or time-aware retrieval strategies, may improve alignment between dynamic disaster scenes and relatively static reference imagery. Efficiency can be further improved by introducing lightweight feature extractors, approximate nearest neighbor search, or hierarchical filtering strategies that progressively narrow down candidate sets. Additional geographic priors, including road networks, POI distributions, or other forms of volunteered geographic information, may provide complementary constraints, especially in cases with highly ambiguous toponyms. It is also worth exploring adaptive strategies that adjust the balance between semantic and visual cues depending on input quality. Integrating uncertainty estimation and human-in-the-loop mechanisms could make the system more transparent and reliable, which is important for high-stakes disaster response tasks where fully automated decisions may not always be sufficient.

\section{Conclusion}

This study proposes DisasterTD, a cross-view toponym disambiguation framework that integrates MLLM-based candidate generation with ViT-enhanced cross-view matching across SMI, SVI, and RSI. Experimental results show that the proposed framework substantially improves disaster geolocalization accuracy under diverse toponym ambiguity scenarios. Across five spatial thresholds (i.e., 1000 m, 500 m, 250 m, 100 m, and 50 m), DisasterTD consistently outperforms all baselines while also reducing mean and median geolocalization errors. DisasterTD achieves geolocalization accuracies of 71.62\% within 1000 m, 62.36\% within 500 m, 57.99\% within 250 m, 52.09\% within 100 m, and 47.01\% within 50 m, with a mean error of 11.33 km and a median error of 0.68 km. Compared with the MLLM-only baseline, it achieves an average improvement of 21.79\%; relative to the direct cross-view matching between SMI and RSI, the gain is 17.22\%; and compared with the joint cross-view matching pipeline among SMI, SVI, and RSI, it still delivers a 10.88\% improvement. These results demonstrate that combining semantic reasoning with multi-view visual evidence provides a more reliable solution for fine-grained disaster geolocalization. Future work will extend the framework to broader disaster scenarios and geographic contexts while exploring more efficient cross-view retrieval strategies for real-time disaster response applications.


{\small
\bibliography{references}
\bibliographystyle{IEEEtranN}}




\vspace{-10pt}

\vfill
\end{document}